# Deep Learning Development Environment in Virtual Reality

Kevin C. VanHorn, Meyer Zinn, & Murat Can Cobanoglu

**Abstract**—Virtual reality (VR) offers immersive visualization and intuitive interaction. We leverage VR to enable any biomedical professional to deploy a deep learning (DL) model for image classification. While DL models can be powerful tools for data analysis, they are also challenging to understand and develop. To make deep learning more accessible and intuitive, we have built a virtual reality-based DL development environment. Within our environment, the user can move tangible objects to construct a neural network only using their hands. Our software automatically translates these configurations into a trainable model and then reports its resulting accuracy on a test dataset in real-time. Furthermore, we have enriched the virtual objects with visualizations of the model's components such that users can achieve insight about the DL models that they are developing. With this approach, we bridge the gap between professionals in different fields of expertise while offering a novel perspective for model analysis and data interaction. We further suggest that techniques of development and visualization in deep learning can benefit by integrating virtual reality.
**Availability:** github.com/Cobanoglu-Lab/VR4DL

**Index Terms**—Virtual reality, Machine learning, Deep learning, Neural nets, Visualization

✦

## 1 INTRODUCTION

DEEP learning (DL) has emerged as a broadly useful field of machine learning that is deployed in many domains. In parallel, technological advancements have made virtual reality (VR) viable in routine applications. VR offers inherent advantages for visualization, which DL can benefit from. To leverage the synergy, we developed a VR platform to build DL models in a more intuitive and immersive environment.

Deep learning is a subset of machine learning that processes data through successive layers between the model input and output. DL has become increasingly relevant in natural language processing, image recognition, and other fields of artificial intelligence [1]. Convolutional neural networks (ConvNets) are a specific class of deep neural networks designed for sequential data. This data can take on various multi-dimensional forms, including audio, video, and visual imagery. In regard to image analysis, ConvNets have become the dominant approach for nearly all recognition and detection tasks and have produced remarkable results for segmentation [1]. Applications include the labeling of pathology images on a per-pixel basis [2], object recognition for self-driving cars [3], and segmentation in medical imaging [4].

Data visualization in deep learning is crucial to model construction, diagnostics, and a general intuition into how deep neural networks function. Proper analysis of neural networks is often clouded by uncertainty due to the large quantities of intermediate layers. To address this limitation, interactive techniques have been developed such as the live observation of layer activations which has been shown to build valuable intuition [5]. Data representations tend to retain pieces of information that resemble important features of the input [6]. Observation of these features can help validate the integrity of a model and identify edge cases that may occur. Research has been dedicated to many such visualization methods. Examples in this field include overlaying heat maps and highlighting features that contribute to the final classification of an image [7]. Hidden features of the data can also be extracted; such techniques have been implemented for a wide range of applications [8], [9], [10], including Deep Car Watcher, which constructs color maps from driving data that can be used to infer a drivers intention [11]. Overall, the visualization of neural networks can offer critical insights that benefit users at all levels of experience.

Virtual reality (VR) offers distinct advantages over conventional data visualization approaches. Techniques for coding, development, and training are much more immersive in VR compared to other relevant mediums. Heightened immersion can increase productivity, retention, and (in our case) ease of understanding. Similarly, VR enables more intuitive user interfaces that bolster the effect of interactivity. As a result, we can use VR to perform tasks more easily and with greater comprehension. We define the functional aspect of virtual reality by two core components: immersion and interactivity [12]. These components have improved in recent years as hardware becomes less cumbersome. For visuals, we continue to achieve higher performance in quality, resolution, and frame rate. Developers are also exploring interaction methods that push a sensory encompassing VR experience. Methods include spatial audio, haptic interfaces, motion platforms [13], and olfactory displays [14]. In this regard, VR will likely be at the forefront of emerging media and technology in the coming decades.

By its nature, VR minimizes outside distractions. Headwear encompasses the user's eyes to reduce "information

• Kevin VanHorn, Meyer Zinn, and Murat Can Cobanoglu were with the Lyda Hill Department of Bioinformatics, University of Texas Southwestern Medical Center, Dallas, Texas, 75390.
E-mail: {kevin.vanhorn, murat.cobanoglu}@utsouthwestern.edu

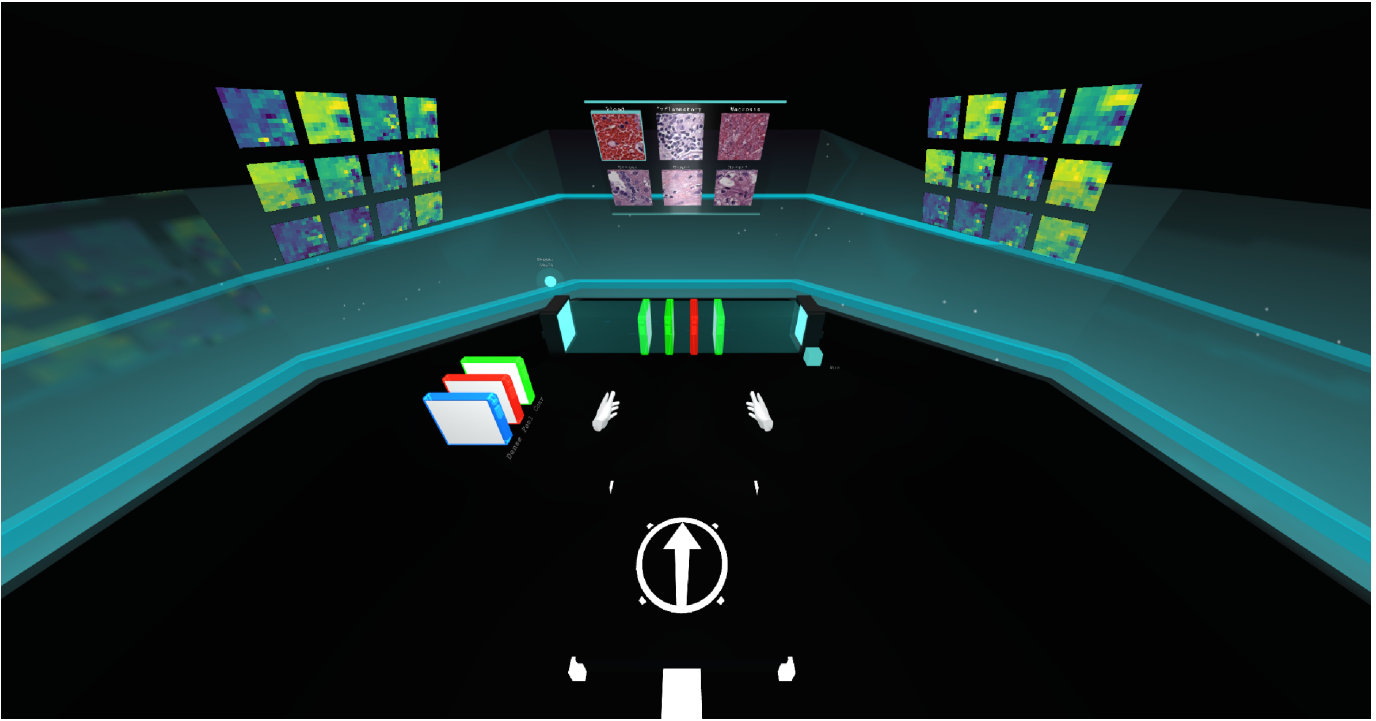

Fig. 1. An overview of our virtual reality environment.

clutter" and other distractions [15]. VR constrains visual feedback and applies varying levels of interactivity to create a sense of presence. We define presence as the feeling that the user is actually there in the environment [16]. Even applications that do not resemble reality can attribute to this phenomenon. By immersion of the senses, we can see benefits in a wide range of fields. Applications vary in design from entertainment and education to psychological therapy [17], [18]. For construction safety training, researchers have seen VR platforms to be more effective than conventional methods for maintaining attention and focus [19]. Both of these benefits are integral to any development environment.

Interactivity in virtual reality acts as a bridge between the application and user. Methods of interaction in VR draw many parallels to actions in real life. VR platforms translate movements from the physical world to the virtual world with high levels of precision. We can take advantage of this by designing controls that mimic physical actions. As an example, a user often navigates the VR camera by turning his/her head and looking around the environment. With this approach, the user can attain positional awareness with ease. We can take advantage of this new paradigm by extending workflows and data representation into the virtual world. Traditional methods of navigating three dimensions on a desktop environment lack functionality. This type of environment impairs navigation and interactivity by separating the user from the action taken. Instead, VR can implement intuitive movements for interaction. An additional restriction of most environments is the inability to represent 3D objects in two dimensions. Virtual reality introduces a method of visualization that mimics day-to-day life. Altogether, VR offers the potential to improve existing workflows and methodologies for data representation.

## 2 RELATED WORK

Various methods exist in two and three dimensions to represent convolutional neural networks. In this domain, we examine algorithmic visualizations and interactive platforms. Visualizations extract data from a model during or after training and are crucial to analysis within interactive platforms. These platforms offer a means with which to construct and/or visualize neural nets by control of various parameters. We propose that a virtual reality platform could complement existing techniques. We also suggest advantages of our VR platform that enable users to construct ConvNets with ease no matter their skillset.

### 2.1 Solutions in 2D

The domain of 2D visual representations for ConvNets is rich with evolving techniques. Each has its own function and purpose over different degrees of granularity. Work in algorithmic representations continues to grow and is beneficial to understanding ConvNets. Many of the platforms that take advantage of these visualizations are limited to a given dataset.

Algorithmic representations display information relevant to the model for understanding and debugging purposes. A user can identify what the computer sees and then adjust the model with the gathered insight. Examples include activation layers, heat maps, filter visualization, feature visualization, and saliency [7]. It is worth noting that researchers need to analyze many of these methods on simple models before extending to deep neural nets [20]. Our solution introduces an interactive environment that can apply evolving representations. To clarify, we encourage the use of a VR platform to better navigate between layers and display valuable information. We can project this information on a 2D surface or present it as a tangible 3D object.

There are many platforms in 2D that offer insight into how ConvNets function. These frameworks offer a method of network visualization limited to some representational extent. Graph visualization is common between many platforms [7]. These node-based systems highlight the relationships between components of a ConvNet. Tools such as TensorFlow Graph Visualizer [21] and CNNVis [10] for neural net simplification exist for those who develop ConvNets but are not heavily integrated into workflows. We argue that an interactive experience in VR can streamline productivity over large volumes of data.

Our product is currently targeted at non-expert users. We define a non-expert as a professional whose primary field of expertise is not in machine learning. A non-expert could also be one with little technical knowledge or one who is familiar with programming but has not been exposed to machine learning. Existing tools for this demographic include Tensorflow Playground [22], ConvNetJS [23], and Keras.js [24]. A majority of techniques targeted at non-experts offer minimal customization of the data set. The ability to adjust model parameters is often restricted or nonexistent in such platforms (with Tensorflow Playground as an exception). Our VR framework allows the user to define the model by specifying a sequence of layers. We train the model and report the accuracy within the application. The ability to construct models allows for interactive visualization with real-time results. This style of interaction environment better fosters insight. We also provide the ability to customize the dataset and model parameters, a feature lacking in most 2D platforms. We argue that by introducing the benefits of virtual reality, we can further improve from traditional 2D techniques.

### 2.2 Solutions in 3D

Interactive three-dimensional platforms are less common and borrow from existing 2D visualization methods. The common visualization strategy is to present a web of expandable layers in 3D. Here we examine TensorSpace.js [25] and an interactive node-link visualization by Harley [26]. TensorSpace.js offers multiple "playgrounds" for interacting with pre-built networks. Both methods display the connection of data between levels of the network. When the user hovers over a pixel, the applications display its connections and resulting size. ConvNets involve variable shapes of data matrices between each layer. By visualizing the effect of each layer type, users can gain valuable insight into how a model functions. However, these 3D methods are limited to non-expert users and would need to be further developed for more advanced use. Both methods analyzed focus on digit recognition or existing layers. In this regard, there are heavy customization and interaction constraints in current 3D platforms. In our solution, we separate training and visualization through the gRPC, remote procedure call framework. This allows for complex analysis in real time. As a result, we can better integrate problems with large datasets in fields like pathology with ease.

The field of DL visualization platforms is widely underdeveloped in terms of 3D tools and techniques. Our application introduces model development that solves a customizable classification task. With our approach, the user can build and test a network for any set of images and gain insight from his/her actions. For a non-expert user, existing 3D visualization platforms overload the user with information. These applications can also be difficult and inconsistent to navigate. If a user is unable to navigate the environment with ease, the tool has less educational value. With a virtual reality framework, we can ease this disconnect. VR users are "placed" into the virtual environment, allowing for unrestricted inspection. With virtual reality, existing techniques can experience significant benefits for interaction and visualization.

## 3 BACKGROUND

Convolutional neural networks process data through many layers of artificial neurons. This process is similar to that of the human brain. When such networks take in data, their neurons have individual responses within each layer. A single layer (e.g. convolution) can have multiple filters of the same type. Layers exist in between the input and output of the model. Data passes through a network sequentially until a final judgment forms on the input. For the task of image classification, this judgment specifies a label for the image. ConvNets produce these labels with a level of uncertainty. We can measure the performance of a DL model by its accuracy and loss. As a model becomes more robust, we want to receive predictions with a tighter level of confidence. In our framework, we implement three types of layers in the context of the TensorFlow/Keras DL library. We limit the number of types for easier construction of effective ConvNets by a non-expert user. These layers are visualized as rectangular objects (figure 2) in our application.

Convolutional (conv) layers define small transformations applied to every part of the image. Each convolution element learns to transform a section of the image into a single value. We 'slide' these elements across the entire input. Every conv layer also contains multiple filters of these convolution elements.

Max pooling (pool) layers group the signal from a very small patch of the image into a single value by taking the largest value. They serve to selectively highlight a dominant signal from the surrounding noise. In the process, they reduce the size of the input.

Dense layers connect every part of the information in their input layer to every part of their output layer. They enable the user to bridge the gap between the many signals in the input image and the small number of classes in the label decision.

## 4 DESIGN

We have constructed a VR visualization development environment for deep convolutional neural network models. In our platform, users can easily build and visualize components of a neural network. We compile models in real-time on the local computer or over the network. For faster results, the application can connect to high-performance computing (HPC) nodes. Our solution offers a shallow learning curve for the construction and implementation of deep convolutional networks. Using this approach, users of varying levels of technical knowledge can obtain insight into the neural network. The current platform communicates the fundamentals of deep learning for classification tasks. It also offers an expandable technique for professional and educational model development.

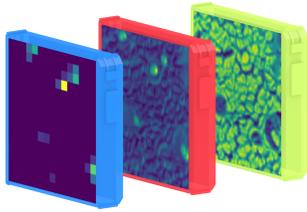

Fig. 2. A render of the representational layer objects in the VR application. Left to right: dense, max pooling, convolutional.

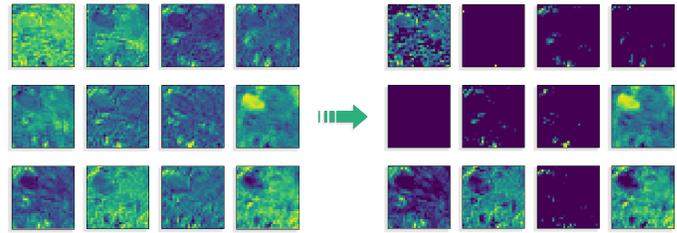

Fig. 3. An example of filters visualized from a convolution layer (left) and the corresponding ReLU activation (right). ReLU activation is applied by default after each conv dense layer.

### 4.1 Environment

The virtual reality environment is simplistic with minimal distractions. The user controls virtual hands to grab, throw, and place objects. The user can additionally point a laser from his/her index finger for item selection. We have designed the environment to function with limited physical movement. The user only needs a small physical space to operate the application and can perform actions while standing or sitting. We minimize possible VR motion sickness with a calm and sleek environment that maintains constant points of reference for the user.

Our VR interface surrounds the user with various tools for building deep learning models. Users can grab layers from a tool shelf to their left and place them within a "workspace." The workspace is a defined region in front of the user that suspends a sequence of layers. Users can insert, remove, and rearrange layers within its bounds. These operations allow for simple construction and modification of a functional ConvNet. After the user grabs a layer from the tool shelf, a new layer of the same type will spawn to fill its place. By this manner, there can be more than one layer of a given type within the model. The user defines the model from left to right and can train it by pressing or selecting a button to the right of the workspace. While computing, a display in front of the user reports the status of training. Upon completion, the same display reports the accuracy of the model against a testing set.

As data passes through a ConvNet, layers reduce the image resolution. In our application, we display the current shape with a peripheral at the top left of the workspace. This dimensionality indicator reflects the state of the model with a representational sphere (figure 4). The shape of the sphere interpolates between sizes to reflect the state of the model. For example, adding a max pooling layer reduces the 128x128 pixel image to a shape of 64x64. For simplicity, we ignore extra dimensions present in the tensor such as filter depth and color in this indicator visualization. Before input passes through a first dense layer, we flatten the 2D image matrix to a 1D array. We display this reduction by flattening the sphere indicator to a line. Without this tool, the rearrangement of layers would have no discernable effect on the model. Since understanding the shape and dimension-ality of the model is crucial to users as they refine a layer sequence, this indicator provides valuable information.

When the model is complete, we report accuracy at the display beneath the workbench. At this point, the application updates layers with their intermediate activations. The image displayed corresponds to the first filter of the respective layer. As the user adds more layers, the resolution of the image reduces to reflect changes in shape. Dense layers after the flatten operation are not visualized because they are not spatially intuitive. We offer a more complete representation of the layer by projecting a 4x3 matrix of such activations at the user's right. This matrix contains additional filters for the layer that the user last touched. We also apply a ReLU activation by default to all convolutional and max pooling layers. The user can view the effects of this activation by holding a button on the controller (figure 3). The user changes the matrix to their right by hovering on or picking up a layer. One can glide his/her hand through the sequence of layers to see a progression of the network. With this technique, the user can see the flow of data as it passes through the ConvNet. The user can then use this insight to modify his/her sequence. For example, the user may see that data is too large throughout the network. At this point, he/she would want to insert max pooling layers before any convolutional layers to trim the data.

Certain models may be excellent at identifying one class of images but not another. Above the workspace, we display image classes that the ConvNet is analyzing. The user can select a class to display activations on for the next build of the model. This information gives context and customization to the classification problem. Rather than stating the task, we display example image patches for the user to observe. When the user selects a class of image, we can visualize activations specific to that class. In this manner, the user can analyze the different effects of the ConvNet on each class. We provide this customization to aid users in the refinement of their models. If many of the intermediate activations are blank, the model may not be effective. In this case, the user can reorder layers to better identify key features of the input image. The ReLU activations can also be insightful to the user. These activations pull out different features depending on the range of intensities in an image class. This may motivate the user to reduce or increase the number of convolutional layers, for example.

### 4.2 User Experience

As discussed in section 1, interactivity is a distinct advantage of a virtual reality environment. We have designed the experience to ease the user into building a DL model. Our primary goal in design was to keep the environment simple and intuitive.

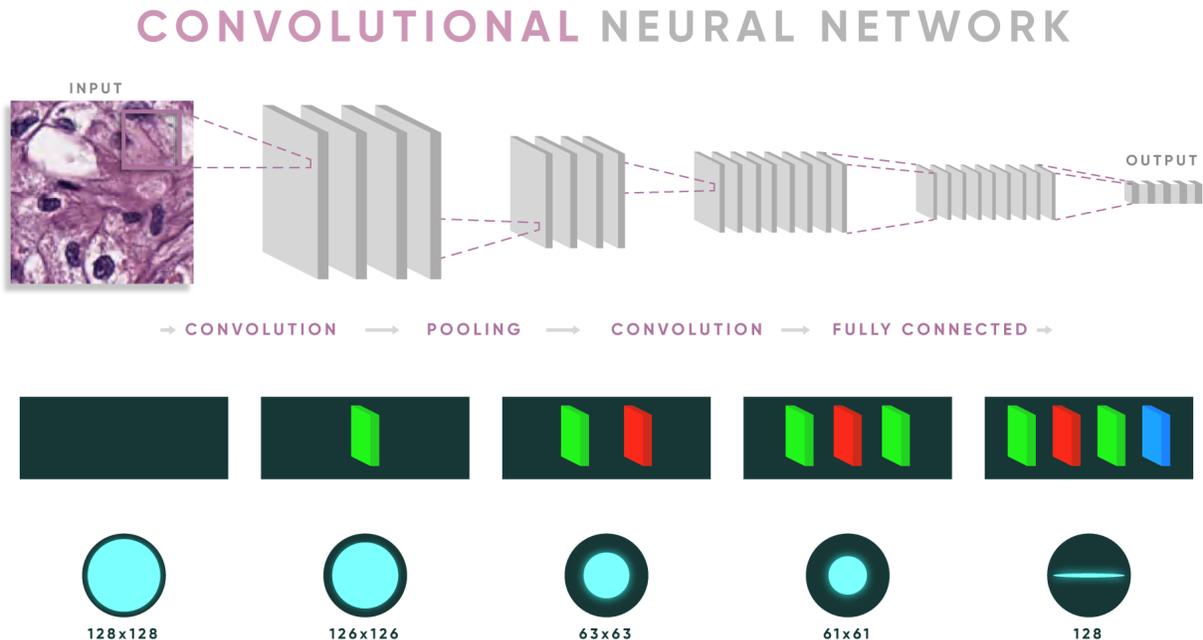

Fig. 4. Traditional representation of a neural network (top), our method of building a model (middle), and our dimensionality indicator within the VR environment to help build intuition as a model is built (bottom).

The core aspect of the application is building the ConvNet. For this functionality, users can grab objects and release them into the workspace. If any arrangement needs to occur, users can again pick up layers and move them to a new location. We keep this functionality simple to ensure that users can interact with the application in a familiar way. An element of familiarity is especially important for users new to virtual reality. The environment will be unfamiliar to all new users, so we design interactions to minimize the learning curve. To discard a layer, the user picks it up and then throws it in any direction. We apply in-game physics to the object at this point so that it falls to the ground and disappears on contact with the floor. In this manner, the user can quickly cast aside irrelevant layers and intuitively modify the model.

Initially, we defined the workspace with a fixed number of translucent containers. In this setup, layers snap to the center of a container when released. However, we observed two drawbacks for this setup: vagueness about whether the translucent objects or the gaps in between are the holders, and an implicit suggestion to fill all the containers to complete the model building task. These were undesirable intuitions. The user should immediately identify the target area. Likewise, smaller models may be more effective thus there should not be a discriminating intuition against them. Additionally, this order made model order definition difficult.

To address these concerns, we modeled the workspace after containers from Google's Material Design [27]. We adopted the "slipstream" design, which is an adaptable list that centers layers within a defined data stream. We apply linear interpolation to translate objects when a user modifies the stream. Through this method, the sequence can move to free up space wherever the user is attempting to place a layer. As a result, the interaction model suggests functionality by design rather than instruction. Furthermore, we can also ensure that restrictions of the ConvNet are upheld. When users release dense layers at any position in the workspace, we propagate them to the back. This communicates a common design principle about this layer type without direct instruction. With this redesign, we make the model building process more clear. Users can quickly append, insert, and remove layers with intuitive grab and release actions.

## 5 Implementation

While deep learning models are useful in many domains, we had to focus on one task to build a functional prototype. The specific task we chose to address is the classification of 128x128 RGB image patches. These patches originate from a larger set of tissue images. To make our prototype serve a useful goal, we chose to focus on the pathology pixel classification problem where DL models are the state-of-the-art classifiers. Googles AI for metastatic breast cancer detection, LYNA, demonstrates a case where deep learning algorithms are faster and more accurate than pathologists [28]. Further study using the same algorithm suggests that by adding the contextual strengths of human experts, we can better employ DL to improve diagnostic speed and accuracy [29]. For our dataset specifically, we first divide patches into seven classifications based on pathologist annotations: tumor, stroma, background, normal, inflammatory, blood, and necrosis. We then construct testing and training sets from this data. Image classes are used for the final model and also for displayed activations.

The software runs as a standalone application operated by the standard requirements of the Oculus Rift VR headset. We chose this hardware for its intuitive controls that mimic grabbing and releasing. Oculus Rift also has a widely available application programming interface (API) for game engines. We developed the frontend of the application in the Unity game development platform (Unity 2018.2.15f1). We built the DL backend with the TensorFlow/Keras framework using Python3. The Unity application and backend communicate core requests via gRPC. gRPC is a remote procedure call system initially developed by Google. It exists as an interface between different programming languages over the network. We render and export images locally using OpenCV. We modified the standard "Oculus Integration for Unity" asset for base interaction functionality. One can configure our software to implement any set of image patches.

### 5.1 Configurability

Users can configure our application to process and display their own custom datasets. We update the in-game visuals and back-end model to reflect changes in the data. In this manner, we provide the benefit of our environment across classification tasks and domains. With further adaptation, our application could construct neural networks for additional fields. One such task is the label-free prediction of fluorescence images for transmitted-light microscopy [30]. In this case, we would not flatten the image, output full-sized image predictions instead.

Our application provides a .config file that allows the user to specify parameters for the model and dataset. Users can modify the batch size, the number of epochs, and details specific to each layer type. If the user provides custom data, he/she can specify the number of classes, the name of each category, and the number of datapoints. The frontend and backend both update to reflect these changes. If the user specifies new classes, the input selector above the workbench will update accordingly. In this situation, the user must provide up to six input images for the generation of activations.

As a proof-of-concept, we have integrated histopathologic image patches from the Camelyon16 dataset that assess the presence of metastatic tissue for breast cancer [28]. We obtain patches for binary classification from the PatchCamelyon benchmark [31]. This adaptation of the Camelyon16 dataset labels image patches based on whether a tumor tissue exists in the center region of an image. With our integration of this new problem set, we have achieved accuracy upwards of 80% with minimal testing. We quickly found that to construct effective models for this dataset, users needed to construct larger networks by adding more layers.

### 5.2 Constraints

Our application automatically enforces the constraints inherent to convolutional neural networks. We eliminate the need to debug errors by ensuring that all models respect inherent constraints of the Keras architecture. This speeds up the initial learning process for experts and non-experts alike. Through this method, we can effectively eliminate the "boilerplate" present in high-level APIs. Users can thereby develop a model for the first time in seconds. We uphold the following constraints to ensure a robust and consistent user experience.

1) We present the user with layers of three types: convolutional, max pooling and dense. Within these types, we can communicate the core DL concepts relevant to ConvNets. Users can build an effective neural network while lowering the complexity and potential to "break" the model.

2) We propagate dense layers to the end of the model because of the accompanying flatten layers effect on the shape of the model. Before the first dense layer, we flatten the model to one dimension. After this point, no multidimensional layers can exist. To ensure that this restriction is upheld, we do not permit the user to place a conv or pool layer after a dense layer.

3) We limit the number of layers in the workspace to 10 for a consistent user experience. As the number of layers increases in a model, so does the compile time and training time. Thus, to avoid the user needing to wait excessive amounts of time, we restrict the model size. This restriction does not necessarily lower the accuracy of the model. Throughout testing, we have seen that models with excessive numbers of layers in our dataset tend to have diminishing returns for our classification task.

4) We implement the core input and output layers of the model beforehand. By doing so, we avoid grossly inaccurate or dysfunctional models. This process includes defining the initial shape with a conv layer and classifying the final result.

5) Models produced within our application are sequential and apply only to classification tasks. Custom datasets must adhere to this constraint. As discussed in section 5.1, we can modify the program to support any number of ConvNet problem domains. For a consistent experience, we have chosen to focus the application datasets such as ours.

## 6 RESULTS

In this section, we analyze the performance of the models constructed within our application. For this purpose, we specify a distinction between exploratory and production methods of interactivity. The first method benefits users looking to explore a dataset and obtain general insight into what models are effective. The second provided more detailed data for intensive analysis, better for refining the model.

### 6.1 Model Training and Evaluation

Our application gives the user a choice of how computationally expensive evaluation should be. By default, we use a small training set for streamlined exploration. In this situation, results often fluctuate on each run, even with the same model. This is due to the sample size and non-deterministic nature of DL. If a user wants more detailed, production-quality information, he/she can select a larger dataset. This will report results truer to the performance of the model, but the process will take more time.

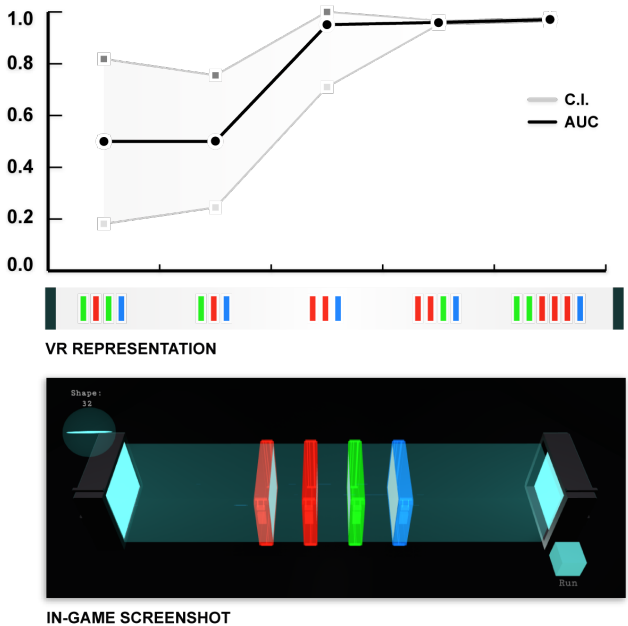

Fig. 5. An example of user progression within the VR application. The x-axis signifies user-defined layer sequences such as the screenshot depicted.

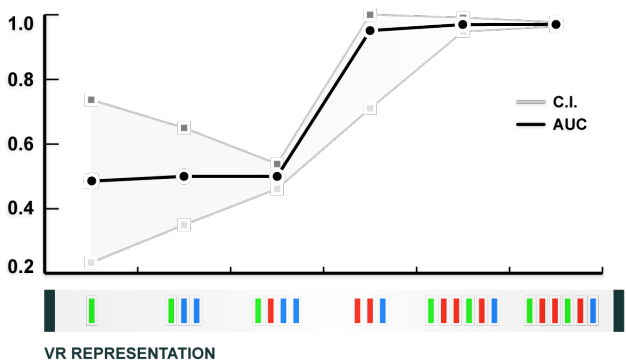

Fig. 6. A second example of user progression. Confidence interval and AUC are reported for each model built (y-axis).

In both modes, we compile the model with the binary cross entropy loss function and adam optimizer in Keras. Conv layers have 32 filters and a kernel size of (3,3). Conv layers are always followed by a ReLU activation as previously mentioned. Max pooling layers have a pool size of (2,2), and dense layers have 128 neurons. Every model has a convolutional layer and its accompanying activation to begin. At the end of a model, a dense layer exists with 7 neurons and a softmax activation. The number of neurons corresponds to the final number of classes and can vary with the dataset. If the user does not insert a dense layer, we put a flatten layer before the final classification. Otherwise, we perform this flatten immediately before the first dense layer that the user provides.

Each model uses 60% of the data for training, 20% for validation, and 20% for testing. We use validation data for early stopping to minimize overfitting. We evaluate the performance of the final model on the testing set.

For the option of rapid training and evaluation, we use a balanced set of 7000 data points. The application compiles and fits models using the training and validation data. We save the best iteration in this process. Here, we stop early if convergence occurs or if the program reaches a set number of epochs. We enforce a maximum to ensure rapid development and testing. Finally, the application reports the accuracy of a model on the testing set to the user. Overall, this model involves fewer epochs, a smaller patience value, and less data. Thus, the accuracy reported to the user is not wholly consistent between runs. This can be a hindrance to more advanced learning and model development. For this use, we recommend the second option.

The second option takes more time-intensive calculations but does not necessarily have to be slow. Using gRPC, we can connect the application to a series of high-performance computing nodes (HPC). For accessibility purposes, a larger dataset is not the default. For this option, we choose to measure the performance of models using a fivefold stratified cross-validation strategy. The application trains models at each fold until convergence with a high patience value. This means, that the model can run over more epochs to achieve more consistent results. We train on a balanced set of 14,000 data points as well. Instead of model accuracy, we report the area under the receiver operating characteristic curve (AUC). This new primary metric of evaluation is performed on a new set of data that is not shown to the model during training. This set is also balanced and consists of 21,000 image patches. To measure the final performance of the model we report the median fivefold mean AUC and its 95% confidence interval (CI) [32]. In the following subsection, we obtain the AUC and CI reported in figure 5 using this strategy.

## 6.2 Progression

There are certain patterns that result in a more successful model. These patterns are specific to the classification domain and features of the dataset. In this section we discuss how users can discern patterns for our dataset through iterative experimentation and informed evaluation. The progression of the user is a term used here to describe the vital process of improving a model for a given problem. Model progression is crucial to understanding and designing an effective DL model. Both experts and non-experts can benefit because each problem and domain poses a unique set of challenges.

Through observation and testing, we have identified certain patterns that our interface can provide to the user. Patterns can be specific to the dataset or general to the domain. General patterns tend to cover aspects inherent to ConvNets. These are more relevant for non-experts to understand how neural nets function. In figure 5, we explore some possible progression strategies. These progressions can communicate various important patterns to the user. As the user progresses, he/she can better refine a model by tightening the CI and increasing the AUC.

In figure 5, the user starts with a larger, underperforming network. He/she reduces the problem to three layers for simplification. By identifying the effectiveness of multiple pooling layers, the user can further progress. The insight here is that max pooling layers reduce the dimensionality of the image so that the model can analyze specific features. By the 5th stage, the user has brought back in conv layers to achieve an AUC of 0.97. With this progression, we communicate the need for pooling layers if conv elements are to be effective. The first and fourth stages differ by only one layer, yet they produce significantly different results. This observation illustrates the volatility of networks and the need for reducing dimensionality.

In figure 6, the user follows a different progression to achieve the same model. Initially, the user builds up from a single conv layer. He/she then identifies an issue and then continues to improve the model. After stage 3, the user is able to identify that multiple dense layers are not necessary for small models. In the final two stages, the user identifies the benefit of grouping layers of the same type.

# 7 DISCUSSION

An interactive VR visualization approach has promising results for cross-domain machine learning applications. For image classification, one can implement this technique throughout the field of bioinformatics. We presented an example, virtual environment for the classification of pathology data. Our application targets new users for rapid model building. However, one can expand this approach to the nth degree. Researchers and pathologists could use one such VR environment for any deep learning task.

## 7.1 Applications

We tailored our tool to users with little to no knowledge of machine learning. We visualize processes that are difficult to understand and interact with otherwise. Our method is useful because it permits the construction of high quality models with minimal effort. For education, teachers can integrate VR applications into curriculum and training at all levels. Such applications can relay vital concepts fundamental to those learning about deep learning. The properties inherent to VR in section 1 could greatly benefit educational environments. By combining evolving technology and traditional curriculum, teachers can boost attentiveness and learning. Virtual reality introduces a novel technique for visual learning. This technique offers a tactile and interactive learning strategy to stimulate students' minds.

Fully-immersive VR environments have immense potential for professional applications. Experienced developers in the field of artificial intelligence are one of such audiences. We can streamline production and better guide visual insight with this technique. Traditional platforms fail to capitalize on advances in immersion and interactivity. Virtual reality offers a new paradigm of development that can implement existing visualizations. We can continue to build on state of the art representations in VR with this approach.

Presence and immersion specific to VR offer many benefits in efficiency and understanding.

Two classes of professionals are already using visualization techniques for DL [33]. The first class, which we have touched on, is model builders/developers. Developers benefit from new methods of development and visualization for debugging purposes. With detailed technical information, this class of users can better construct DL models. These models are then deployed to solve real-world problems. The second class is that of the target users. Target users do not need to understand the details a model but need to interact with it on some level. Convenient visualization is especially important for this class to communicate relevant information. A pathologist using a DL model to aid in labeling regions of an image is an example of this class.

## 7.2 Future Work

Here we discuss some of the identifiable pitfalls and areas of improvement in our work. We stress that one can apply the technique proposed with any level of complexity. Technical improvements and customization using modern DL visualizations benefit different target audiences.

Customization is the primary concern that one would need to address in future work. Our application sacrifices customization for fast and easy model implementation. It lacks in the ability of the user to fine tune more advanced parameters. This could include the optimization technique and more extensive layer properties. These additions benefit more advanced users but can overwhelm those new to DL. Without more sophisticated parameter testing, fine-tuned control also slows development. In a professional setting, companies can provide these tools while interfacing with VR. We chose to build the product around fundamental features but recognize potential improvements. For this purpose, we propose that an application of this type be modular. In this setup, developers hide advanced features from the user but allow for extensive customization.

Context is an additional aspect that one would need to expand on. New users are often confused by the field of deep learning and the purpose of the program. For this purpose, we recorded an introductory video to explain the controls and problem domain. Even with further explanation, many users were unsure what problem they were solving. Pathology and image classification techniques are extensive fields that can confuse users. To this effect, we see the benefit of further improvement to ease understanding for new users.

Visualization techniques are a strong point of potential improvement. The solution would benefit greatly from more sophisticated tools that provide analysis in 2D and 3D. Many of the algorithms and frameworks in section 2 would be useful here. These visualizations heighten the potential for insight and provide multi-faceted analysis. Expanding layers and 3D visualization of data flow would be particularly helpful.

# 8 CONCLUSION

We have discussed the benefits of our tool for building DL models in VR and highlighted further improvements.

We analyzed existing strategies for data representation and platforms that employ them. We argued for the extension of visualization to virtual reality. In this regard, educational and professional environments could achieve heightened efficiency and insight. To illustrate a simple use case, we built our application to classify tumor image patches from two different datasets. We found our approach to be effective and robust for building ConvNets with ease.

## ACKNOWLEDGMENTS

Tumor images for classification were provided by Drs. Satwik Rajaram and Payal Kapur who is funded by the Kidney cancer SPORE grant (P50CA196516). The software is a derivative of work from the UT Southwestern hackathon, U-HACK Med 2018, and has continued development under the same Principal Investigator (Murat Can Cobanoglu) and lead developer (Kevin VanHorn). The project was originally proposed by Murat Can Cobanoglu, with the final code being submitted to the NCBI-Hackathons GitHub under the MIT License. Hackathon was sponsored by BICF from funding provided by Cancer Prevention and Research Institute of Texas (RP150596). We would like to thank hackathon contributors Xiaoxian Jing (Southern Methodist University), Siddharth Agarwal (University of Texas Arlington), and Michael Dannuzio (University of Texas at Dallas) for their initial work in design and development.

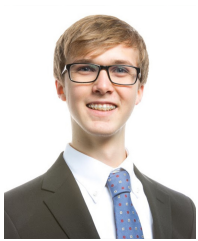

**Kevin VanHorn** Kevin VanHorn received his B.S. in Computer Science at the University of Texas at Dallas and is currently pursuing a M.S. at the same institution. For his undergraduate education, he was awarded the Terry Scholarship. He is currently developing visualization solutions for deep learning to aid in bioinformatics research at UT Southwestern Medical Center. Kevin has experience as a graphic artist and independent game developer.

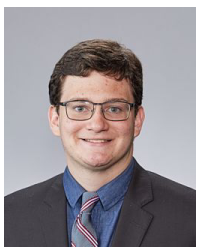

**Meyer Zinn** Meyer Zinn is a rising senior at St. Mark's School of Texas with an interest in computer science and artificial intelligence. He was awarded the Rice University Book Award, the Engineering Sciences Award, and the University of Rochester Xerox Award for Innovation and Information Technology. Previously, he developed a real time location tracking system for the Medical Artificial Intelligence and Automation lab's smart clinic project in the Department of Radiation Oncology. He has participated in several hackathons and competitions, and his team was the runner-up at EarthxHack 2019.

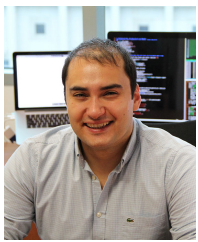

**Murat Can Cobanoglu** Dr. Cobanoglu received his undergraduate training in computer science (major) and mathematics (minor) from Sabanc University in stanbul, where he was invited as a recipient of the university's most competitive scholarship. He completed his M.S. in computer science in Sabanci University, supported by a fellowship from the Scientific and Technological Research Council of Turkey. He attended the Carnegie Mellon University University of Pittsburgh Joint Ph.D. Program in Computational Biology. He joined the Lyda Hill Department of Bioinformatics as the inaugural UTSW Distinguished Fellow. He works on integrating machine learning into drug discovery.